\documentclass[journal]{IEEEtran}

\ifCLASSINFOpdf
\else
\fi

\usepackage{amsmath,amssymb,amsfonts}
\usepackage{algorithmic}
\usepackage{bm}
\usepackage{booktabs} 
\usepackage{multirow} 
\usepackage{graphicx}
\usepackage{pgfplots}
\usepackage{float}
\usepackage{hyperref}
\pgfplotsset{compat=1.17} 
\usepgfplotslibrary{groupplots}
\usepackage{caption}
\usepackage{subcaption}
\usepackage{cite}
\usepackage{tabularx}
\usepackage{adjustbox}

\begin{document}

\title{Flexible and Explainable Graph Analysis for EEG-based Alzheimer's Disease Classification}

\author{Jing Wang, Jun-En Ding, Feng Liu, Elisa Kallioniemi, Shuqiang Wang, Wen-Xiang Tsai, Albert C. Yang

\thanks{J. Wang, J. Ding and F. Liu are with the Department of Systems and Enterprises at Stevens Institute of Technology in Hoboken, NJ 07030, United States, and they are also with Semcer Center for Healthcare Innovation at Stevens Institute of Technology. Corresponding author: F. Liu, Email: fliu22@stevens.edu.}
\thanks{E. Kallioniemi is with New Jersey Institute of Technology, Newark, NJ, United States.}
\thanks{S. Wang is with Shenzhen Institutes of Advanced Technology, Chinese Academy of Sciences, Shenzhen, 518055, China}
\thanks{A. Yang and W. Tsai are with the Institute of Brain Science, College of Medicine, National Yang-Ming Chiao Tung University, Taiwan.}
\thanks{Research reported in this study was partially supported by NINDS of NIH under Award Number R21NS135482. The content is solely the authors' responsibility and does not necessarily represent the official views of the NIH.}
}

\maketitle

\vspace{-5mm}
\begin{abstract}
Alzheimer's Disease is a progressive neurological disorder that is one of the most common forms of dementia. It leads to a decline in memory, reasoning ability, and behavior, especially in older people. The cause of Alzheimer's Disease is still under exploration and there is no all-inclusive theory that can explain the pathologies in each individual patient. Nevertheless, early intervention has been found to be effective in managing symptoms and slowing down the disease's progression.
Recent research has utilized electroencephalography (EEG) data to identify biomarkers that distinguish Alzheimer's Disease patients from healthy individuals. Prior studies have used various machine learning methods, including deep learning and graph neural networks, to examine electroencephalography-based signals for identifying Alzheimer's Disease patients. In our research, we proposed a Flexible and Explainable Gated Graph Convolutional Network (GGCN) with Multi-Objective Tree-Structured Parzen Estimator (MOTPE) hyperparameter tuning. This provides a flexible solution that efficiently identifies the optimal number of GGCN blocks to achieve the optimized precision, specificity, and recall outcomes, as well as the optimized area under the Receiver Operating Characteristic (AUC).
Our findings demonstrated a high efficacy with an over 0.9 Receiver Operating Characteristic score, alongside precision, specificity, and recall scores in distinguishing health control with Alzheimer's Disease patients in Moderate to Severe Dementia using the power spectrum density (PSD) of electroencephalography signals across various frequency bands. Moreover, our research enhanced the interpretability of the embedded adjacency matrices, revealing connectivity differences in frontal and parietal brain regions between Alzheimer's patients and healthy individuals.

\end{abstract}

\begin{IEEEkeywords}
EEG, Alzheimer's Disease, Dementia, Graph Neural Network, Explainability
\end{IEEEkeywords}


\section{Introduction}
%
%
%
%
\IEEEPARstart
{D}{ementia} is a group of syndromes characterized by brain impairments such as memory loss, declined thinking abilities, and limited reasoning, which interfere with an individual's daily functioning~\cite{dementia2019dementia}.
More than 25 million people have dementia with Alzheimer's Disease, accounting for 75\% of the cases~\cite{qiu2009epidemiology}.
The occurrence of Alzheimer's Disease is closely associated with age; with the world population aging, it is estimated that by 2050, the prevalence will quadruple, resulting in 1 in 85 persons will be living with the disease, with about 43\% requiring a high level of care~\cite{brookmeyer2007forecasting}. In Taiwan, Alzheimer's Disease is the leading cause of dementia among the elderly population, with a prevalence of approximately 1.7 - 4.3\%~\cite{fuh2008dementia}. Therefore, early diagnosis and intervention can be of great benefit to society.
While the exact cause of Alzheimer's disease (AD) have yet to be fully understood~\cite{li2018dementia}, research suggests that age, female gender, low educational attainment, and prior head injuries are among the risk factors that contribute to its development~\cite{abeysinghe2020alzheimer}. Early intervention at the earliest stages of the disease can significantly reduce healthcare expenses and enhance the patient's overall quality of life~\cite{dekosky2003early}. As such, timely diagnosis is paramount in assisting those affected by AD.

The way we diagnose and manage Alzheimer's disease has undergone a transformation from relying solely on clinical symptom reporting that are Alzheimer's related brain dysfunctions to a more accurate diagnosis method that combines clinical evaluation with AD pathology, including bodily fluids and imaging studies with good specificity~\cite{weller2018current}. 

Modern clinical studies employ a variety of imaging techniques, including electrobiological measurements like electroencephalography (EEG) and magnetoencephalography (MEG), which present data through a parameter graph over time~\cite{hussain2022modern}. Physical principle-based techniques such as computer tomography (CT), magnetic resonance imaging (MRI), functional MRI (fMRI), positron emission tomography (PET), and single-photon emission computed tomography (SPECT) are also frequently utilized.
In the field of brain research, extensive machine learning and deep learning algorithms have been applied to EEG-based data in recent years as EEG is a non-invasive way to read the electrical signals generated by brain structures~\cite{teplan2002fundamentals}, making it an effective tool to understand the brain activities affected by AD. It has two main advantages. The first is characteristic of the electrical recording system, which is high precision time measurements where the fast-changing electrical activity in the brain can be recorded, and the second is a non-invasive procedure that allows researchers to have access to HC's brain eeg data~\cite{zion2007eeg}.
Identifying AD from HC using direct EEG signals is challenging because of the variability among subjects, which arises from anatomical and physiological differences~\cite{salazar2019facing}. Consequently, numerous researchers have utilized advanced algorithms to overcome this problem.


\section{Related Work}
To identify the AD of different stages with Healthy Control (HC) or to classify AD of different states, various Machine Learning approaches have been involved to examine brain patterns through EEG data analysis including but not limited to K-nearest-neighbor (KNN), Linear Discriminant Analysis (LDA) and Support Vector Machine (SVM), among these methods, SVM is mostly used one~\cite{modir2023systematic}. 
Tait et al. employed an SVM predictor to differentiate AD, healthy older adults (HOA), and mild cognitive impairment (MCI) patients based on microstate analysis, achieving a sensitivity and specificity of more than 80\%~\cite{tait2020eeg}. Miltiadous et el. demonstrated that SVM attained over 90\% accuracy, sensitivity, and specificity in distinguishing AD with HC and frontotemporal dementia (FTD) with HC~\cite{miltiadous2021alzheimer}. Trinh et al. utilized SVM on the extracted task-induced intra-subject spectral power variability of resting-state EEGs features and achieved 74\% to 80\% accuracy in distinguishing AD with HC, MCI with HC, and AD with MCI~\cite{trinh2021identifying}. Hsiao et al. utilized a conformal kernel-based fuzzy support vector machine (CKF-SVM) to circumvent the overfitting of outliers that EEG features often encountered due to intra-subject and inter-subject variations~\cite{hsiao2021eeg}. 

Researchers have developed various deep learning techniques to identify Alzheimer's disease (AD) through EEG signals. One approach was proposed by Morabito and colleagues, who utilized a Convolutional Neural Network (CNN) to detect hidden patterns and distinguish mild cognitive impairment (MCI) with AD, achieving an average sensitivity and specificity of 80\%~\cite{morabito2016deep}. Similarly, Zhao and his team employed a Deep Belief Network (DBN) to extract learning features in an unsupervised manner, which were then sent into SVM to classify AD with HC with an accuracy of 92\%~\cite{zhao2015deep}. Kim et al. utilized a	deep Multi-layer perceptron neural network (MLPNN) using the relative power (RP) and proved that deep neural network (DNN) enhances the performance of MCI and HC detection~\cite{kim2018detection}.

In recent years, researchers have incorporated graph theory analysis to understand the interconnectivity between different pathological processes linked to Alzheimer's disease~\cite{miraglia2022brain, stam2007graph}. This idea is grounded in the nature of graph neural networks, where nodes remain invariant, therefore it extends the Convolutional Neural Network (CNN) to the Non-Euclidean Space. In this way it best preserves the brain's characteristics when analyzing connections between regions.
Shan et al. proposed a new method for classifying healthy controls (HC) and Alzheimer's disease (AD) patients in an eyes-closed (EC) state using EEG signals. Their method, the EEG-based Spatial-temporal graph convolutional network (STGCN), combines the adjacency matrix of functional connectivity between EEG channels with the dynamics of signals among each channel, providing a more comprehensive analysis. The STGCN achieved a classification accuracy of 92.3\%, which is better than the state-of-the-art methods~\cite{shan2022spatial}.
Demir et al. proposed the Graph Attention Networks for EEG signals (EEG-GAT) that extends the EEGNet by designing an interpretable graph model via the multi-head attention mechanism to learn the connection between different regions in the brain~\cite{demir2022eeg}.
Klepl and his team have developed a GCN model called the adaptive gated graph convolutional network (AGGCN). This model first enhanced the Power Spectrum Density (PSD) features through the use of a one-dimensional convolutional neural network (1D CNN). And then, they incorporated a Gated Graph Convolutional Network (GGCN) Encoder, which selectively retains important information at each scale instead of integrating the entire neighbourhood into the node embedding. Additionally, they implemented an adaptive structure-aware pooling (ASAP) mechanism that identifies the most important clusters of nodes, which are then passed into a fully connected layer for classification. This innovative approach ensures that the model generates consistent explanations of its predictions~\cite{klepl2023adaptive}.

Researchers are continuously working on improving the performance of classification models while also focusing on making them more explainable. A study by Khare et al. has introduced an automated adaptive and explainable system for detecting Alzheimer's disease (Adazd-Net) using EEG signals. The system utilizes SHapley Additive exPlanations (SHAP) and Local Interpretable Model-agnostic Explanations (LIME) to interpret the classification model based on the adaptive flexible analytic wavelet transform~\cite{khare2023adazd}.

\section{Data and Preprocessing}
\subsection{Participants}
The study recruited participants from the Dementia Clinic at the Neurological Institute, Taipei Veterans General Hospital in Taiwan~\cite{yang2013cognitive}. The data set included 108 patients with Alzheimer's Disease (AD) and 15 healthy control participants (HC), ranging in age from 46 to 95, with an average age of 77 years.
Using the Clinical Dementia Rating (CDR) scale, participants were categorized by the severity of their condition, with a score of 0 indicating no dementia, 0.5 indicating very mild, 1 indicating mild, and 2 or higher indicating severe dementia.
In a previous study using the same dataset, Liu et al.~\cite{liu2021phenotyping} analyzed patients with mild Alzheimer's disease who had a CDR rating of 1, as well as healthy controls (HC). Their research showed that EEG power correlates with behavioral and psychological symptoms. Our study aims to distinguish between AD patients and HC controls using the same dementia severity criteria and to find biomarkers that correlate with mild AD. Therefore, we separately analyzed AD patients with a CDR rating of 0.5, 1, and 2 with HC controls.
The AD group with very mild dementia consisted of 15 subjects, with an average age of 78 (SD 10.27), including 7 females and 8 males. The AD group with mild dementia consisted of 69 subjects, with an average age of 78 years (SD 6.75), including 42 females and 27 males. The AD group with Moderate to Severe Dementia consisted of 24 subjects with an average age of 78 (SD 11.92), including 10 females and 14 males. The HC group included 15 subjects, with an average age of 69.87 years (SD 9.55), including 9 females and 6 males. See Table~\ref{tab:demographics}.

\begin{table*}[h!]
\centering
\caption{Demographic Characteristics of AD and Test Groups}
\label{tab:demographics}
\adjustbox{max width=\textwidth}{\begin{tabular}{l|l|l|l|l}
\hline
Group & Age & Gender & Education & MMSE \\
\hline
Test Group (N=15) & 69.87 ± 9.55 & F/M 9/6 & 14.07 ± 8.83 & 28.67 ± 1.05 \\
Very Mild Dementia AD Group (N=15) & 77.67 ± 10.27 & F/M 7/8 & 10.80 ± 4.55 & 24.13 ± 4.14 \\
Mild Dementia AD Group (N=69) & 78.00 ± 6.75 & F/M 42/27 & 7.90 ± 5.03 & 18.91 ± 5.20 \\
Moderate to Severe Dementia AD Group (N=24) & 78.17 ± 11.92 & F/M 10/14 & 7.67 ± 5.83 & 12.08 ± 5.03 \\
\hline
\end{tabular}
}
\end{table*}

\subsection{Signals}

\begin{figure}[!h] 
  \centering
  \includegraphics[scale=0.23]{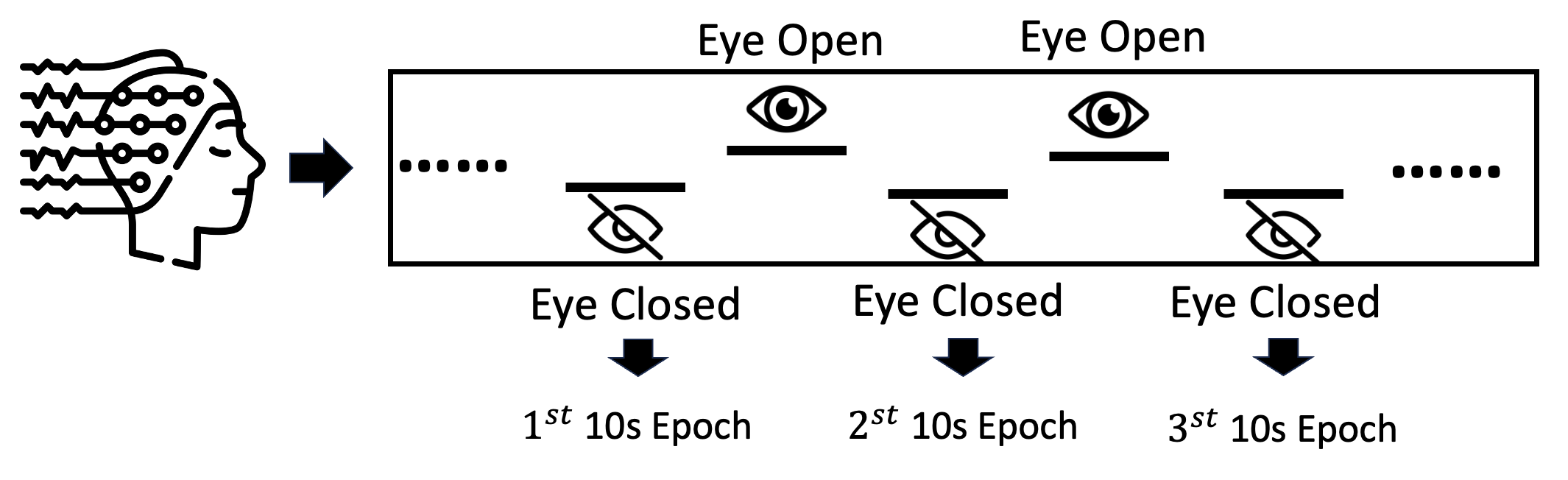}
  \caption{Workflow}
  \label{fig:test_flow}
\end{figure}

Each patient began with a 5 minute habituation to the examining environment, then three separate 10-20s sessions of eye resting state EEG signals from the nineteen electrodes (Fp1, Fp2, F7, F3, Fz, F4, F8, T3, C3, Cz, C4, T4, T5, P3, Pz, P4, T6, O1, and O2) were collected according to international 10-20 system at a sampling rate of 256Hz as shown in Figure~\ref{fig:test_flow}.

The initial filter settings were a low-pass filter of 70 Hz, a high-pass filter of 0.05 Hz, a notch filter of 60 Hz, and electrode impedances below 3 k$\omega$.
After conducting a visual inspection, we removed the bad channels T6 and O2, resulting in 17 electrodes' data for further analysis.
The acquired recordings were then fed into Independent Component Analysis (ICA) to remove the artifacts, thus leading to cleaner signals that could be further processed.

The spectrum power density(PSD) represents the distribution of EEG time series power distribution over frequency and is often used to provide insights into the abnormalities of the brain associated with AD~\cite{wang2015power}. 
We divided the frequency bands in our study into four bins, namely beta (13-40 Hz), alpha (8-13 Hz), theta (4-8 Hz), and delta ($<$4 Hz). To analyze the data, we broke down the epoch into overlapping windows, with a window size of 5 seconds and a step size of 0.5 seconds. The multi-taper method with DPSS tapers~\cite{slepian1978prolate} was used to calculate the power spectrum density for the four EEG bands.

\section{Methods}
\begin{figure*}[h] 
  \centering
  \includegraphics[width=\textwidth]{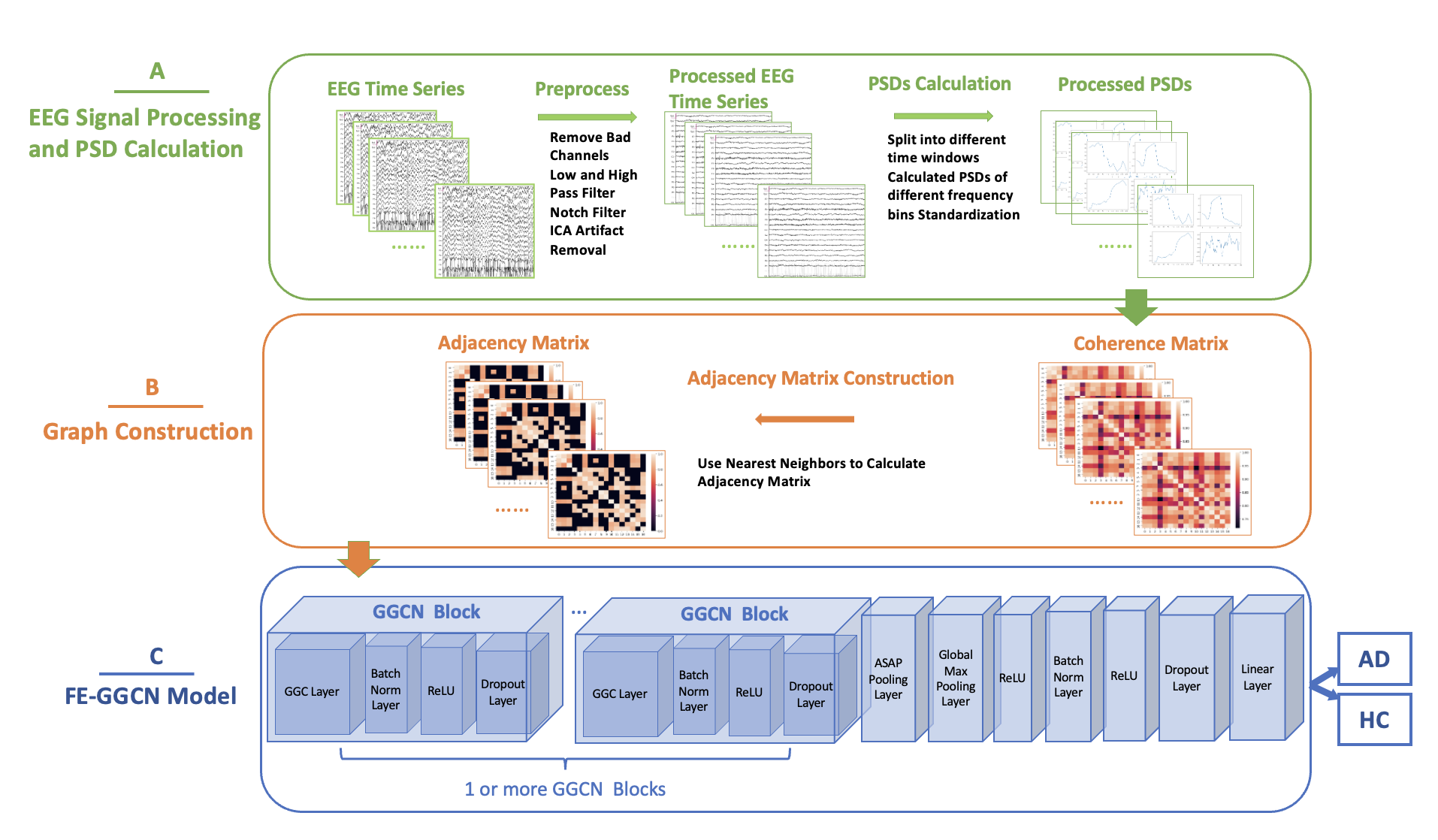}
  \caption{Architecture of the Workflow. (A) Preprocessing of EEG signals, which involves eliminating bad channels, artifact removal via ICA, and computing the Power Spectral Density (PSD) for each frequency band. (B) Construction of graphs begins with the computation of the functional connectivity matrix between channels, followed by the application of the Nearest Neighbors algorithm to form the adjacency matrix. (C) Implementation of the Gated Graph Convolutional Network (GGCN) module, followed by graph pooling and fully connected layers for classification purposes.}
  \label{fig:Architecture of the Workflow}
\end{figure*}

The modeling process consists of a few steps. Firstly, the EEG time series undergoes preprocessing, cleaning, and calculation of Power Spectral Densities (PSDs) for each time window and band. After that, we standardize the PSDs and create a graph using Phase Lag Index (PLI) and Phase Locking Value (PLV) and an unsupervised Nearest Neighbors algorithm. The resulting graph is then utilized for classification through the Graph Convolutional Network and optimized parameters are determined through the use of hyperparameter tuning techniques as shown in Figure~\ref{fig:Architecture of the Workflow}.

\subsection{Graph Construction}
To obtain the graph representation based on the PSDs, we followed a three-step process. Firstly, we standardized the Power Spectral Densities (PSDs) of each node using Z-scores. This was done to ensure that the PSDs were transformed into a common scale and that the resulting correlation matrix would not be biased toward any specific node. Subsequently, we calculated the connectivity matrix \(C \in \mathbb{R}^{N \times N} \) with \( N = 17 \) by computing the Phase Lag Index (PLI) ~\cite{stam2007phase} and Phase Locking Value (PLV) ~\cite{lachaux1999measuring} between each pair of nodes. This step allowed us to determine the strength and direction of the relationship between each pair of nodes. Lastly, we applied Unsupervised Nearest Neighbors Learning to the connectivity matrix. This algorithm identifies the nearest nodes to a given node based on the precomputed distance. As a higher correlation indicates a lower distance, we then created a sparse graph representation based on the coherence matrix. This graph representation retains only the strongest correlations, effectively filtering out weak and noisy connections. The resulting Adjacency Matrix \(A \in \mathbb{R}^{N \times N} \) with \( N = 17 \) can be used to analyze the network properties of the brain and to identify important nodes or hubs within the network.
\begin{equation}
\text{PLI} = \left| \mathbb{E}\left[ \text{sign}\left( \text{Im}(S_{xy}) \right) \right] \right|
\end{equation}

\begin{equation}
\text{PLV} = \left| \mathbb{E}\left[ \frac{S_{xy}}{|S_{xy}|} \right] \right|
\end{equation}

\subsection{Gated Graph Convolutional Network Classification Module}
\subsubsection{Gated Graph Convolutional Network Block}
The Graph Neural Network (GNN) represents an evolved form of the Convolutional Neural Network (CNN), specifically engineered to process graph-based data within non-Euclidean spaces. This type of data often involves topographical structures that traditional approaches may struggle to handle effectively. In the context of neuroscience, brain channels are an example of graph-based data, making graph-based models particularly useful for identifying brain disorders. In this study, we employed one or more blocks of the Gated Graph Convolutional Network (GGCN)~\cite{li2015gated} for the task of EEG-based dementia classification. 

GGCN addresses the limitations found in traditional GCN, particularly when propagating information across long distances. In GCNs, all information from these neighborhoods is incorporated into the embeddings, which can be problematic as the information is aggregated blindly in the brain graphs without regulation. GGCN resolves this issue by using a gated mechanism with gated recurrent units (GRU)~\cite{cho2014properties} to selectively determine which information each spatial scale should retain. This allows for more precise control over the information flow during the propagation process, ensuring that only relevant data is emphasized and retained in the embeddings~\cite {klepl2023adaptive}.

The input to the GGCN classifier is a graph defined as \( G = (V, A, D_{\text{PSD}}) \), where \( V \) represents the set of nodes, with \( N=|V| \), \( A \) represents the set of edges which is the adjacency matrix learned, and \( D_{\text{PSD}} \) represents the set of features which is the standardized PSD features.
The message passing of the GGCN graph is denoted as:
\begin{equation}
\mathbf{m}_i^{(l+1)} = \sum_{j \in \mathcal{N}(i)} e_{j,i} \cdot \mathbf{\Theta} \cdot \mathbf{h}_j^{(l)}
\end{equation}
\begin{equation}
\mathbf{h}_i^{(l+1)} = \textrm{GRU} (\mathbf{m}_i^{(l+1)},
\mathbf{h}_i^{(l)})
\end{equation}
where the $\mathbf{h}_j^{(l)}$ is the node features, $e_{j,i}$ is the edge weight, $\mathbf{\Theta}$ is the learnable weight matrix, and $\mathbf{m}_i^{(l+1)}$ is the message from neighboring nodes. 

After the node embeddings were computed, a batch normalization layer was applied to them. This layer adjusts the mean and standard deviation of each batch, which reduces the impact of internal covariate shift and helps the model converge faster. Next, a Rectified Linear Unit (ReLU) activation layer was used to introduce non-linearity in the output of the batch normalization layer. Finally, to prevent overfitting, a dropout layer was added, which randomly drops out a fraction of the nodes during training, forcing the model to learn more robust features.

\subsubsection{Node Pooling}
Two layers of node pooling were implemented to the node embeddings. The first layer involved using an adaptive structure aware pooling (ASAP)~\cite{ranjan2020asap} technique, which is a novel sparse hierarchical pooling method. This method enabled learning the subgraph information hierarchically, which ultimately led to learning better global features with improved edge connectivity in the pooled graph.
The ASAP approach first considers each node as the medoid of a cluster, capturing local neighbor information within a fixed radius of h-hops, which helps to capture the information in the graph sub-structure. So the cluster assignment matrix is denoted as \( S_{i, j} \) representing the membership of node \(v_i\in V\) in cluster \(c_h(v_j)\) maintains the sparsity of the cluster assignment matrix \(S\) akin to the original graph's adjacency matrix \(A\).
Then, it utilizes a new variant of a self-attention mechanism called Msater2Token (M2T)~\cite{ranjan2020asap} to learn the overall representation of a cluster by attending to relevant nodes, thus determining the cluster assignment matrix \(S\).
The master query of the initial cluster embedding is:
\begin{equation}
    m = \max_{j \in N(i)} x_j
\end{equation}
where the \(N(i)\) is the neighbors of node \(i\). 
The attention score is calculated as:
\begin{equation}
    \alpha_{i, j} = softmax(\omega^T\sigma(Wm_i \mathbin\Vert x_j))
\end{equation}
where \(\omega^T\) is a learnable vector and \(W\) is a learnable matrix. The attention score \(\alpha_{i, j}\) represents the membership strength of node \(v_{j}\) in cluster \(c_h(v_i)\) which can be used as the cluster assignment matrix. Thus, the cluster assignment matrix \(S\) is defined as:
\begin{equation}
    S_{i,j} = a_{i,j}
\end{equation}
Therefore the cluster representation is defined as:
\begin{equation}
    x_{i}^{c} = \sum_{j=1}^{c_{h}(v_{i})}\alpha_{i,j}x_{j}
\end{equation}
The cluster embedding used the local extrema convolution (LEConv)~\cite{ranjan2020asap} which computes the cluster fitness score \(\phi_{i}\) as:
\begin{equation}
    \phi_{i} = \sigma(x_{i}^{c}\Theta_{1}+\sum_{j \in N(i)}A_{i,j}^{c}(x_{i}^{c}\Theta_{2}-x_{j}^{c}\Theta_{3}) )
\end{equation}
where \(\Theta_{1}\), \(\Theta_{2}\), and \(\Theta_{3}\) are learnable parameters, \(\sigma\) is the activation function. And the cluster embedding is calculated as: \(\hat{X^{c}}=\Phi \odot X^{c}\), where \(\Phi\) is the cluster fitness vector, \(\odot\) is the element-wise multiplication, \(X^{c}\) is the initial cluster representation.
The TopK~\cite{gao2019graph} was then used to rank the fitness score, the selected top \(\left \lceil kN \right \rceil\) fitness score's indices is denoted as:
\begin{equation}
   \hat{X}^{c}=TOP_{k}(\hat{X}_{c}, \left \lceil kN \right \rceil)
\end{equation}
where k is the pooling ratio. The pruned cluster assignment matrix \( \hat{S}\in R^{N\times\left \lceil kN \right \rceil} \).  The new adjacency matrix is calculated as $A^{p} = \hat{S}^{T} \hat{A}^{c} \hat{S}$, which will then be used to identify biomarkers on the scalp that can differentiate the AD and HC groups. 

The second layer of pooling used a global max pooling technique, which works by aggregating the node features across each graph in the batch:
\begin{equation}
\mathbf{r}_i = \mathrm{max}_{n=1}^{N_i} \, \mathbf{x}_n
\end{equation}
This enabled the creation of a single vector representation for each graph, condensing node-level information into a graph-level representation.

\subsubsection{Fully Connected Classifier Layer}
At last, the nodes are fed into the fully connected classifier by applying a ReLU non-linearity, introducing regularization with dropout, and finally mapping to the output classes with a linear transformation. The Cross-Entropy Loss was used as the loss function for the model training.

\subsection{Hyperparameter Optimization Framework}
During our hyperparameter tuning process, we utilized a novel Bayesian-based optimization approach called the multiobjective tree-structured Parzen estimator (MOTPE)~\cite{ozaki2022multiobjective}. This approach is particularly effective for models or datasets that are computationally expensive. It is an extension of the widely-used Tree-Structured Parzen Estimator (TPE)~\cite{bergstra2011algorithms}, which has been widely used in recent research. The TPE algorithm categorizes hyperparameters into "good" and "bad" groups based on a performance threshold. It then uses kernel density estimation to model these distributions, with subsequent sampling focused on areas where the ratio of the density of the "good" distribution to the combined densities of both "good" and "bad" distributions is the highest. This method improves the optimization process's efficiency and effectiveness by prioritizing promising areas and reducing the need for random or exhaustive searches, directing the search towards regions of the hyperparameter space that are more likely to yield improvements.

The hyperparameter tuning has been accomplished through the utilization of the Optuna Hyperparameter Optimization Framework~\cite{akiba2019optuna}. This framework operates by specifying a multi-objective function that accepts hyperparameters as input and produces a score indicating the model's performance acquired via those hyperparameters. 

\subsection{Baseline Models}
The SVM model which utilized the Power Spectral Densities (PSDs) from each band is used as baseline model. This model follows the same training, validation, and testing splits as the proposed model, ensuring that the evaluation and reporting of metrics for the baseline models are consistent with those of the proposed model, thus enhancing the reliability of the results. The processing approach involved standardizing the data, applying Principal Component Analysis (PCA) for dimensionality reduction, and using a Support Vector Machine (SVM) classifier to distinguish between HC and AD.

We employed the grid search technique to identify the optimal hyperparameters. These hyperparameters include:
\begin{itemize}
    \item Number of components for PCA: $\text{pca\_\_n\_components}$ with values $0.95$, $0.90$, $0.85$
    \item Regularization strength: $\text{svm\_\_C}$ with values $0.1$, $1$, $10$, $100$
    \item Kernel coefficient: $\text{svm\_\_gamma}$ with values $1$, $0.1$, $0.01$, $0.001$
    \item SVM kernel type: $\text{svm\_\_kernel}$ with options 'rbf' and 'sigmoid'
\end{itemize}

\subsection{Implementation and Evaluation}
The EEG signals were preprocessed using MNE package~\cite{gramfort2014mne}, the Graph Learning was conducted with the aid of Scikit-Learn~\cite{pedregosa2011scikit}, the graph neural network was implemented using Pytorch~\cite{paszke2019pytorch} and PyTorch Geometric~\cite{fey2019fast}, and the hyperparameter tuning was performed with Optuna~\cite{akiba2019optuna}.

The dataset was divided into training, validation, and test sets in a 3:1:1 ratio. Initially, the 5-time repeated 5-fold stratified cross-validation method was used to separate the test set from the training and validation sets. Subsequently, each training-validation set was further divided into training and validation sets using the first split of the stratified K-fold method with K=4. This resulted in 25 iterations of the cross-validation for both validation and testing purposes. By preserving the proportion of AD and HC subjects in the entire dataset, this method ensures that the evaluation of the model's performance is both robust and consistent across different data subsets. 



We used Adam~\cite{kingma2014adam} as the optimizer, and we implemented early stopping where if the validation loss does not decrease for 15 epochs it will stop training to prevent overfitting.

\begin{table*}[!h]
\centering
\caption{Performance Comparison of Models and Features for AD in Very Mild Dementia}
\label{tab:performance_comparison_0_1}
\begin{tabular}{lllccccccc}
\toprule
\multirow{2}{*}{Model} & \multirow{2}{*}{Feature} & \multirow{2}{*}{Segment} & \multirow{2}{*}{Band} & \multicolumn{6}{c}{Metrics} \\
\cmidrule(lr){5-10}
 &  &  &  & AUC & F1 & Precision & Recall & Accuracy & Specificity \\
\midrule
\multirow{12}{*}{Flex GGCN} & \multirow{4}{*}{PLI} & \multirow{4}{*}{SEG01} & Delta & 0.724 ± 0.179 & 0.663 ± 0.197 & 0.767 ± 0.339 & 0.493 ± 0.314 & 0.7 ± 0.163 & 0.907 ± 0.150 \\ 
 &  &  & Theta & 0.82 ± 0.206 & 0.741 ± 0.200 & 0.84 ± 0.328 & 0.587 ± 0.287 & 0.767 ± 0.163 & \textbf{0.947} ± 0.122 \\ 
 &  &  & \textbf{Alpha} & \textbf{0.976} ± 0.070 & \textbf{0.839} ± 0.155 & \textbf{0.842} ± 0.164 & \textbf{0.947} ± 0.122 & \textbf{0.853} ± 0.128 & 0.76 ± 0.275 \\ 
 &  &  & Beta & 0.702 ± 0.162 & 0.706 ± 0.153 & 0.675 ± 0.138 & 0.88 ± 0.186 & 0.72 ± 0.147 & 0.56 ± 0.205 \\ 
 & \multirow{4}{*}{PLI} & \multirow{4}{*}{SEG02} & Delta & 0.756 ± 0.222 & 0.701 ± 0.191 & 0.717 ± 0.217 & 0.72 ± 0.244 & 0.707 ± 0.190 & 0.693 ± 0.229 \\ 
 &  &  & Theta & 0.76 ± 0.255 & 0.686 ± 0.212 & 0.695 ± 0.348 & 0.64 ± 0.364 & 0.72 ± 0.175 & 0.8 ± 0.231 \\ 
 &  &  & Alpha & 0.629 ± 0.182 & 0.577 ± 0.160 & 0.65 ± 0.314 & 0.467 ± 0.267 & 0.613 ± 0.131 & 0.76 ± 0.241 \\ 
 &  &  & \textbf{Beta} & \textbf{0.911} ± 0.130 & \textbf{0.895} ± 0.156 & \textbf{0.936} ± 0.142 & \textbf{0.893} ± 0.155 & \textbf{0.9} ± 0.141 & \textbf{0.907} ± 0.222 \\ 
 & \multirow{4}{*}{PLI} & \multirow{4}{*}{SEG03} & Delta & 0.787 ± 0.204 & 0.649 ± 0.217 & 0.713 ± 0.390 & 0.48 ± 0.341 & 0.693 ± 0.174 & 0.907 ± 0.150 \\ 
 &  &  & Theta & \textbf{0.967} ± 0.077 & 0.923 ± 0.102 & 0.97 ± 0.081 & 0.893 ± 0.182 & 0.927 ± 0.095 & 0.96 ± 0.108 \\ 
 &  &  & Alpha & 0.907 ± 0.124 & 0.813 ± 0.160 & 0.897 ± 0.226 & 0.72 ± 0.261 & 0.827 ± 0.137 & 0.933 ± 0.133 \\ 
 &  &  & \textbf{Beta} & \textbf{0.967} ± 0.097 & \textbf{0.937} ± 0.099 & \textbf{0.98} ± 0.068 & \textbf{0.907} ± 0.177 & \textbf{0.94} ± 0.093 & \textbf{0.973} ± 0.090 \\ 
\midrule
\multirow{12}{*}{Flex GGCN} & \multirow{4}{*}{PLV} & \multirow{4}{*}{SEG01} & Delta & 0.618 ± 0.222 & 0.592 ± 0.211 & 0.613 ± 0.241 & 0.627 ± 0.255 & 0.607 ± 0.199 & 0.587 ± 0.271 \\ 
 &  &  & Theta & 0.904 ± 0.200 & 0.805 ± 0.186 & \textbf{0.864} ± 0.226 & 0.773 ± 0.262 & 0.813 ± 0.178 & \textbf{0.853} ± 0.212 \\ 
 &  &  & Alpha & \textbf{0.967} ± 0.059 & \textbf{0.876} ± 0.105 & 0.851 ± 0.136 & \textbf{0.96} ± 0.108 & \textbf{0.88} ± 0.100 & 0.8 ± 0.189 \\ 
 &  &  & Beta & 0.789 ± 0.240 & 0.75 ± 0.153 & 0.755 ± 0.181 & 0.893 ± 0.155 & 0.767 ± 0.141 & 0.64 ± 0.282 \\ 
 & \multirow{4}{*}{PLV} & \multirow{4}{*}{SEG02} & Delta & 0.527 ± 0.213 & 0.483 ± 0.187 & 0.522 ± 0.129 & 0.773 ± 0.278 & 0.54 ± 0.165 & 0.307 ± 0.248 \\ 
 &  &  & Theta & 0.533 ± 0.267 & 0.648 ± 0.170 & 0.703 ± 0.248 & 0.587 ± 0.271 & 0.667 ± 0.156 & 0.747 ± 0.217 \\ 
 &  &  & Alpha & 0.847 ± 0.122 & 0.751 ± 0.173 & 0.913 ± 0.227 & 0.587 ± 0.271 & 0.773 ± 0.148 & \textbf{0.96} ± 0.108 \\ 
 &  &  & Beta & \textbf{0.898} ± 0.191 & \textbf{0.875} ± 0.179 & \textbf{0.92} ± 0.220 & \textbf{0.827} ± 0.269 & \textbf{0.887} ± 0.154 & 0.947 ± 0.154 \\ 
 & \multirow{4}{*}{PLV} & \multirow{4}{*}{SEG03} & Delta & 0.802 ± 0.157 & 0.744 ± 0.154 & 0.755 ± 0.177 & 0.8 ± 0.231 & 0.753 ± 0.150 & 0.707 ± 0.217 \\ 
 &  &  & Theta & 0.887 ± 0.132 & 0.799 ± 0.192 & 0.782 ± 0.180 & \textbf{0.933} ± 0.133 & 0.813 ± 0.172 & 0.693 ± 0.282 \\ 
 &  &  & Alpha & \textbf{0.947} ± 0.118 & \textbf{0.877} ± 0.132 & 0.883 ± 0.151 & 0.907 ± 0.150 & \textbf{0.88} ± 0.129 & 0.853 ± 0.190 \\ 
 &  &  & Beta & 0.938 ± 0.118 & 0.846 ± 0.203 & \textbf{0.92} ± 0.271 & 0.733 ± 0.327 & 0.867 ± 0.163 & \textbf{1.0} ± 0.000 \\ 
\midrule
\multirow{12}{*}{SVM} & \multirow{4}{*}{PSD} & \multirow{4}{*}{SEG01} & Delta & 0.538 ± 0.245 & 0.387 ± 0.227 & 0.366 ± 0.199 & 0.44 ± 0.309 & 0.393 ± 0.169 & 0.347 ± 0.240 \\ 
 &  &  & Theta & 0.496 ± 0.275 & 0.39 ± 0.308 & 0.413 ± 0.337 & 0.4 ± 0.340 & \textbf{0.493} ± 0.208 & \textbf{0.587} ± 0.271 \\ 
 &  &  & Alpha & 0.538 ± 0.180 & \textbf{0.444} ± 0.238 & \textbf{0.488} ± 0.286 & 0.467 ± 0.313 & 0.487 ± 0.188 & 0.507 ± 0.300 \\ 
 &  &  & Beta & \textbf{0.544} ± 0.183 & 0.408 ± 0.226 & 0.387 ± 0.221 & \textbf{0.48} ± 0.328 & 0.427 ± 0.134 & 0.373 ± 0.272 \\ 
 & \multirow{4}{*}{PSD} & \multirow{4}{*}{SEG02} & Delta & \textbf{0.538} ± 0.275 & 0.43 ± 0.305 & 0.437 ± 0.322 & 0.467 ± 0.365 & 0.487 ± 0.226 & 0.507 ± 0.285 \\ 
 &  &  & Theta & 0.513 ± 0.271 & 0.412 ± 0.261 & 0.456 ± 0.312 & 0.413 ± 0.287 & \textbf{0.513} ± 0.169 & \textbf{0.613} ± 0.322 \\ 
 &  &  & Alpha & 0.46 ± 0.216 & 0.486 ± 0.278 & 0.411 ± 0.231 & \textbf{0.613} ± 0.373 & 0.493 ± 0.153 & 0.373 ± 0.272 \\ 
 &  &  & Beta & 0.449 ± 0.165 & \textbf{0.497} ± 0.193 & \textbf{0.471} ± 0.198 & 0.587 ± 0.287 & 0.46 ± 0.158 & 0.333 ± 0.298 \\ 
 & \multirow{4}{*}{PSD} & \multirow{4}{*}{SEG03} & Delta & 0.433 ± 0.240 & \textbf{0.582} ± 0.248 & 0.527 ± 0.220 & \textbf{0.68} ± 0.319 & \textbf{0.573} ± 0.195 & 0.467 ± 0.298 \\ 
 &  &  & Theta & \textbf{0.549} ± 0.249 & 0.415 ± 0.237 & 0.391 ± 0.244 & 0.48 ± 0.299 & 0.42 ± 0.164 & 0.36 ± 0.297 \\ 
 &  &  & Alpha & 0.416 ± 0.175 & 0.562 ± 0.173 & \textbf{0.665} ± 0.273 & 0.573 ± 0.241 & 0.56 ± 0.210 & 0.547 ± 0.410 \\ 
 &  &  & Beta & 0.52 ± 0.205 & 0.501 ± 0.227 & 0.578 ± 0.292 & 0.52 ± 0.314 & 0.553 ± 0.154 & \textbf{0.587} ± 0.287 \\ 
\midrule
\bottomrule
\end{tabular}
\end{table*}

\begin{table*}[!h]
\centering
\caption{Performance Comparison of Models and Features for AD in Mild Dementia}
\label{tab:performance_comparison_0_2}
\adjustbox{max width=\textwidth}{\begin{tabular}{lllccccccc}
\toprule
\multirow{2}{*}{Model} & \multirow{2}{*}{Feature} & \multirow{2}{*}{Segment} & \multirow{2}{*}{Band} & \multicolumn{6}{c}{Metrics} \\
\cmidrule(lr){5-10}
 &  &  &  & AUC & F1 & Precision & Recall & Accuracy & Specificity \\
\midrule
\multirow{12}{*}{Flex GGCN} & \multirow{4}{*}{PLI} & \multirow{4}{*}{SEG01} & Delta & 0.551 ± 0.191 & 0.709 ± 0.064 & 0.823 ± 0.029 & 0.836 ± 0.106 & 0.717 ± 0.084 & 0.173 ± 0.167 \\ 
 &  &  & \textbf{Theta} & \textbf{0.914} ± 0.124 & \textbf{0.913} ± 0.070 & \textbf{0.955} ± 0.056 & 0.948 ± 0.060 & \textbf{0.917} ± 0.063 & \textbf{0.773} ± 0.294 \\ 
 &  &  & Alpha & 0.909 ± 0.092 & 0.859 ± 0.084 & 0.945 ± 0.056 & 0.871 ± 0.090 & 0.851 ± 0.092 & 0.76 ± 0.241 \\ 
 &  &  & Beta & 0.886 ± 0.098 & 0.874 ± 0.074 & 0.918 ± 0.057 & \textbf{0.951} ± 0.063 & 0.886 ± 0.059 & 0.587 ± 0.302 \\ 
 & \multirow{4}{*}{PLI} & \multirow{4}{*}{SEG02} & Delta & 0.76 ± 0.150 & 0.817 ± 0.084 & 0.885 ± 0.057 & 0.919 ± 0.089 & 0.831 ± 0.081 & 0.427 ± 0.306 \\ 
 &  &  & Theta & 0.713 ± 0.112 & 0.774 ± 0.063 & 0.838 ± 0.032 & \textbf{0.986} ± 0.035 & 0.831 ± 0.046 & 0.12 ± 0.186 \\ 
 &  &  & Alpha & \textbf{0.893} ± 0.104 & 0.798 ± 0.063 & \textbf{0.948} ± 0.053 & 0.772 ± 0.064 & 0.776 ± 0.071 & \textbf{0.8} ± 0.211 \\ 
 &  &  & Beta & 0.889 ± 0.116 & \textbf{0.88} ± 0.063 & 0.909 ± 0.044 & 0.965 ± 0.046 & \textbf{0.891} ± 0.056 & 0.547 ± 0.229 \\ 
 & \multirow{4}{*}{PLI} & \multirow{4}{*}{SEG03} & Delta & 0.93 ± 0.103 & 0.919 ± 0.071 & \textbf{0.959} ± 0.051 & 0.945 ± 0.062 & 0.919 ± 0.071 & \textbf{0.8} ± 0.249 \\ 
 &  &  & Theta & 0.867 ± 0.141 & 0.83 ± 0.078 & 0.937 ± 0.063 & 0.852 ± 0.097 & 0.826 ± 0.082 & 0.707 ± 0.317 \\ 
 &  &  & \textbf{Alpha} & \textbf{0.961} ± 0.082 & \textbf{0.935} ± 0.070 & 0.945 ± 0.053 & \textbf{0.991} ± 0.031 & \textbf{0.943} ± 0.059 & 0.72 ± 0.278 \\ 
 &  &  & Beta & 0.797 ± 0.113 & 0.768 ± 0.045 & 0.837 ± 0.027 & 0.988 ± 0.033 & 0.831 ± 0.022 & 0.107 ± 0.182 \\ 
\midrule
\multirow{12}{*}{Flex GGCN} & \multirow{4}{*}{PLV} & \multirow{4}{*}{SEG01} & Delta & 0.738 ± 0.196 & 0.636 ± 0.160 & 0.655 ± 0.140 & 0.867 ± 0.231 & 0.673 ± 0.129 & 0.48 ± 0.268 \\ 
 &  &  & Theta & 0.6 ± 0.206 & 0.463 ± 0.193 & 0.563 ± 0.143 & 0.933 ± 0.133 & 0.567 ± 0.141 & 0.2 ± 0.298 \\ 
 &  &  & Alpha & \textbf{0.842} ± 0.109 & \textbf{0.896} ± 0.068 & \textbf{0.918} ± 0.048 & \textbf{0.977} ± 0.039 & \textbf{0.907} ± 0.058 & \textbf{0.587} ± 0.254 \\ 
 &  &  & Beta & 0.776 ± 0.189 & 0.831 ± 0.092 & 0.878 ± 0.060 & 0.971 ± 0.042 & 0.861 ± 0.073 & 0.36 ± 0.326 \\ 
 & \multirow{4}{*}{PLV} & \multirow{4}{*}{SEG02} & Delta & 0.538 ± 0.223 & 0.439 ± 0.192 & 0.458 ± 0.248 & 0.52 ± 0.341 & 0.48 ± 0.185 & 0.44 ± 0.294 \\ 
 &  &  & \textbf{Theta} & \textbf{0.951} ± 0.100 & \textbf{0.901} ± 0.143 & \textbf{0.944} ± 0.135 & \textbf{0.893} ± 0.182 & \textbf{0.907} ± 0.134 & 0.92 ± 0.195 \\ 
 &  &  & Alpha & 0.647 ± 0.200 & 0.569 ± 0.169 & 0.677 ± 0.415 & 0.333 ± 0.249 & 0.627 ± 0.127 & 0.92 ± 0.142 \\ 
 &  &  & Beta & 0.902 ± 0.153 & 0.827 ± 0.168 & 0.927 ± 0.222 & 0.707 ± 0.255 & 0.84 ± 0.145 & \textbf{0.973} ± 0.090 \\ 
 & \multirow{4}{*}{PLV} & \multirow{4}{*}{SEG03} & Delta & 0.78 ± 0.166 & 0.834 ± 0.070 & 0.873 ± 0.039 & 0.98 ± 0.038 & 0.864 ± 0.051 & 0.333 ± 0.231 \\ 
 &  &  & Theta & 0.513 ± 0.163 & 0.785 ± 0.069 & 0.846 ± 0.036 & 0.977 ± 0.044 & 0.833 ± 0.053 & 0.173 ± 0.213 \\ 
 &  &  & Alpha & 0.803 ± 0.175 & \textbf{0.876} ± 0.077 & \textbf{0.903} ± 0.054 & \textbf{0.985} ± 0.037 & \textbf{0.897} ± 0.055 & \textbf{0.493} ± 0.300 \\ 
 &  &  & Beta & \textbf{0.814} ± 0.118 & 0.769 ± 0.051 & 0.836 ± 0.024 & \textbf{0.985} ± 0.037 & 0.828 ± 0.036 & 0.107 ± 0.155 \\ 
\midrule
\multirow{12}{*}{SVM} & \multirow{4}{*}{PSD} & \multirow{4}{*}{SEG01} & Delta & 0.539 ± 0.176 & 0.403 ± 0.348 & 0.47 ± 0.386 & 0.372 ± 0.349 & 0.4 ± 0.216 & 0.533 ± 0.432 \\ 
 &  &  & Theta & \textbf{0.657} ± 0.230 & 0.698 ± 0.167 & \textbf{0.913} ± 0.083 & 0.598 ± 0.200 & 0.615 ± 0.155 & \textbf{0.693} ± 0.297 \\ 
 &  &  & Alpha & 0.541 ± 0.167 & 0.563 ± 0.344 & 0.661 ± 0.374 & 0.516 ± 0.338 & 0.536 ± 0.237 & 0.627 ± 0.288 \\ 
 &  &  & Beta & 0.634 ± 0.184 & \textbf{0.749} ± 0.122 & 0.868 ± 0.079 & \textbf{0.682} ± 0.182 & \textbf{0.65} ± 0.138 & 0.507 ± 0.300 \\ 
 & \multirow{4}{*}{PSD} & \multirow{4}{*}{SEG02} & Delta & 0.525 ± 0.191 & 0.585 ± 0.240 & 0.727 ± 0.280 & \textbf{0.514} ± 0.254 & 0.503 ± 0.160 & 0.453 ± 0.399 \\ 
 &  &  & Theta & 0.422 ± 0.177 & 0.408 ± 0.344 & 0.553 ± 0.422 & 0.354 ± 0.326 & 0.416 ± 0.225 & 0.707 ± 0.357 \\ 
 &  &  & Alpha & 0.438 ± 0.158 & 0.49 ± 0.357 & 0.549 ± 0.378 & 0.462 ± 0.365 & 0.47 ± 0.238 & 0.507 ± 0.390 \\ 
 &  &  & Beta & \textbf{0.571} ± 0.169 & \textbf{0.591} ± 0.239 & \textbf{0.831} ± 0.254 & 0.488 ± 0.247 & \textbf{0.532} ± 0.184 & \textbf{0.733} ± 0.211 \\ 
 & \multirow{4}{*}{PSD} & \multirow{4}{*}{SEG03} & Delta & 0.52 ± 0.156 & 0.515 ± 0.311 & 0.596 ± 0.344 & 0.463 ± 0.294 & 0.459 ± 0.197 & 0.44 ± 0.374 \\ 
 &  &  & Theta & 0.532 ± 0.181 & 0.649 ± 0.161 & 0.836 ± 0.109 & 0.559 ± 0.207 & 0.547 ± 0.159 & 0.493 ± 0.314 \\ 
 &  &  & Alpha & 0.662 ± 0.208 & 0.657 ± 0.216 & 0.848 ± 0.264 & 0.558 ± 0.222 & 0.586 ± 0.153 & \textbf{0.72} ± 0.361 \\ 
 &  &  & Beta & \textbf{0.704} ± 0.181 & \textbf{0.753} ± 0.179 & \textbf{0.857} ± 0.192 & \textbf{0.683} ± 0.190 & \textbf{0.669} ± 0.151 & 0.6 ± 0.340 \\ 
\midrule
\bottomrule
\end{tabular}
}
\end{table*}

\begin{table*}[!h]
\centering
\caption{Performance Comparison of Models and Features for AD in Moderate to Severe Dementia}
\label{tab:performance_comparison_0_3}
\adjustbox{max width=\textwidth}{\begin{tabular}{lllccccccc}
\toprule
\multirow{2}{*}{Model} & \multirow{2}{*}{Feature} & \multirow{2}{*}{Segment} & \multirow{2}{*}{Band} & \multicolumn{6}{c}{Metrics} \\
\cmidrule(lr){5-10}
 &  &  &  & AUC & F1 & Precision & Recall & Accuracy & Specificity \\
\midrule
\multirow{12}{*}{Flex GGCN} & \multirow{4}{*}{PLI} & \multirow{4}{*}{SEG01} & Delta & 0.674 ± 0.199 & 0.605 ± 0.157 & 0.688 ± 0.136 & 0.792 ± 0.217 & 0.64 ± 0.149 & 0.4 ± 0.283 \\ 
 &  &  & Theta & \textbf{0.979} ± 0.041 & \textbf{0.959} ± 0.069 & \textbf{0.972} ± 0.064 & 0.968 ± 0.073 & \textbf{0.96} ± 0.068 & \textbf{0.947} ± 0.122 \\ 
 &  &  & Alpha & 0.868 ± 0.133 & 0.912 ± 0.128 & 0.91 ± 0.114 & \textbf{0.992} ± 0.039 & 0.922 ± 0.101 & 0.813 ± 0.251 \\ 
 &  &  & Beta & 0.971 ± 0.062 & 0.927 ± 0.098 & 0.959 ± 0.088 & 0.934 ± 0.112 & 0.929 ± 0.095 & 0.92 ± 0.171 \\ 
 & \multirow{4}{*}{PLI} & \multirow{4}{*}{SEG02} & Delta & 0.974 ± 0.073 & 0.904 ± 0.122 & 0.932 ± 0.103 & 0.934 ± 0.149 & 0.909 ± 0.115 & 0.867 ± 0.211 \\ 
 &  &  & Theta & \textbf{0.992} ± 0.029 & 0.949 ± 0.080 & \textbf{0.979} ± 0.058 & 0.942 ± 0.109 & 0.949 ± 0.080 & \textbf{0.96} ± 0.108 \\ 
 &  &  & Alpha & 0.98 ± 0.048 & \textbf{0.953} ± 0.073 & 0.964 ± 0.072 & 0.966 ± 0.078 & \textbf{0.954} ± 0.071 & 0.933 ± 0.133 \\ 
 &  &  & Beta & 0.871 ± 0.136 & 0.844 ± 0.156 & 0.844 ± 0.127 & \textbf{0.992} ± 0.039 & 0.866 ± 0.121 & 0.667 ± 0.298 \\ 
 & \multirow{4}{*}{PLI} & \multirow{4}{*}{SEG03} & Delta & 0.846 ± 0.147 & 0.801 ± 0.174 & 0.834 ± 0.142 & 0.952 ± 0.102 & 0.829 ± 0.133 & 0.64 ± 0.339 \\ 
 &  &  & Theta & \textbf{0.952} ± 0.081 & 0.888 ± 0.118 & 0.896 ± 0.112 & 0.952 ± 0.085 & 0.894 ± 0.110 & 0.8 ± 0.231 \\ 
 &  &  & Alpha & 0.85 ± 0.104 & 0.79 ± 0.133 & \textbf{0.99} ± 0.049 & 0.674 ± 0.191 & 0.795 ± 0.123 & \textbf{0.987} ± 0.065 \\ 
 &  &  & Beta & 0.888 ± 0.118 & \textbf{0.898} ± 0.104 & 0.899 ± 0.101 & \textbf{0.956} ± 0.089 & \textbf{0.902} ± 0.100 & 0.813 ± 0.190 \\ 
\midrule
\multirow{12}{*}{Flex GGCN} & \multirow{4}{*}{PLV} & \multirow{4}{*}{SEG01} & Delta & 0.732 ± 0.221 & 0.625 ± 0.248 & 0.684 ± 0.366 & 0.59 ± 0.353 & 0.671 ± 0.188 & 0.8 ± 0.267 \\ 
 &  &  & Theta & 0.969 ± 0.044 & 0.923 ± 0.090 & 0.909 ± 0.099 & \textbf{1.0} ± 0.000 & 0.929 ± 0.080 & 0.813 ± 0.212 \\ 
 &  &  & \textbf{Alpha} & \textbf{0.996} ± 0.014 & 0.963 ± 0.060 & 0.959 ± 0.074 & 0.992 ± 0.039 & 0.964 ± 0.057 & 0.92 ± 0.142 \\ 
 &  &  & Beta & 0.991 ± 0.024 & \textbf{0.984} ± 0.044 & \textbf{1.0} ± 0.000 & 0.972 ± 0.076 & \textbf{0.984} ± 0.045 & \textbf{1.0} ± 0.000 \\ 
 & \multirow{4}{*}{PLV} & \multirow{4}{*}{SEG02} & Delta & 0.959 ± 0.068 & 0.907 ± 0.092 & 0.964 ± 0.072 & 0.892 ± 0.143 & 0.908 ± 0.091 & 0.933 ± 0.133 \\ 
 &  &  & Theta & 0.995 ± 0.026 & 0.964 ± 0.084 & 0.99 ± 0.049 & 0.948 ± 0.109 & 0.964 ± 0.085 & 0.987 ± 0.065 \\ 
 &  &  & \textbf{Alpha} & \textbf{0.997} ± 0.013 & \textbf{0.984} ± 0.042 & \textbf{1.0} ± 0.000 & \textbf{0.974} ± 0.071 & \textbf{0.984} ± 0.043 & \textbf{1.0} ± 0.000 \\ 
 &  &  & Beta & 0.961 ± 0.110 & 0.962 ± 0.082 & 0.981 ± 0.067 & 0.964 ± 0.083 & 0.964 ± 0.077 & 0.96 ± 0.144 \\ 
 & \multirow{4}{*}{PLV} & \multirow{4}{*}{SEG03} & Delta & 0.927 ± 0.111 & 0.869 ± 0.134 & 0.915 ± 0.118 & 0.884 ± 0.143 & 0.871 ± 0.131 & 0.853 ± 0.212 \\ 
 &  &  & Theta & 0.869 ± 0.140 & 0.818 ± 0.123 & 0.968 ± 0.073 & 0.74 ± 0.181 & 0.821 ± 0.114 & 0.947 ± 0.122 \\ 
 &  &  & Alpha & 0.976 ± 0.057 & 0.94 ± 0.080 & 0.934 ± 0.091 & \textbf{0.99} ± 0.049 & 0.944 ± 0.073 & 0.867 ± 0.189 \\ 
 &  &  & \textbf{Beta} & \textbf{0.979} ± 0.062 & \textbf{0.963} ± 0.069 & \textbf{0.985} ± 0.050 & 0.956 ± 0.089 & \textbf{0.964} ± 0.068 & \textbf{0.973} ± 0.090 \\ 
\midrule
\multirow{12}{*}{SVM} & \multirow{4}{*}{PSD} & \multirow{4}{*}{SEG01} & Delta & 0.513 ± 0.215 & 0.615 ± 0.152 & 0.732 ± 0.197 & 0.6 ± 0.240 & 0.572 ± 0.153 & 0.52 ± 0.401 \\ 
 &  &  & Theta & \textbf{0.709} ± 0.297 & \textbf{0.765} ± 0.114 & \textbf{0.813} ± 0.164 & 0.766 ± 0.183 & \textbf{0.716} ± 0.130 & \textbf{0.64} ± 0.388 \\ 
 &  &  & Alpha & 0.45 ± 0.260 & 0.749 ± 0.055 & 0.67 ± 0.130 & \textbf{0.902} ± 0.153 & 0.631 ± 0.083 & 0.2 ± 0.340 \\ 
 &  &  & Beta & 0.409 ± 0.261 & 0.702 ± 0.116 & 0.671 ± 0.121 & 0.804 ± 0.229 & 0.609 ± 0.094 & 0.293 ± 0.344 \\ 
 & \multirow{4}{*}{PSD} & \multirow{4}{*}{SEG02} & Delta & 0.551 ± 0.301 & 0.698 ± 0.172 & 0.755 ± 0.192 & 0.694 ± 0.232 & 0.651 ± 0.179 & 0.587 ± 0.356 \\ 
 &  &  & Theta & \textbf{0.781} ± 0.263 & 0.783 ± 0.115 & \textbf{0.88} ± 0.147 & 0.75 ± 0.196 & 0.755 ± 0.125 & 0.76 ± 0.334 \\ 
 &  &  & Alpha & 0.469 ± 0.255 & 0.635 ± 0.154 & 0.646 ± 0.151 & 0.712 ± 0.282 & 0.553 ± 0.117 & 0.307 ± 0.352 \\ 
 &  &  & Beta & 0.695 ± 0.328 & \textbf{0.785} ± 0.157 & 0.878 ± 0.145 & \textbf{0.756} ± 0.213 & \textbf{0.764} ± 0.149 & \textbf{0.773} ± 0.323 \\ 
 & \multirow{4}{*}{PSD} & \multirow{4}{*}{SEG03} & Delta & 0.513 ± 0.212 & 0.583 ± 0.183 & 0.688 ± 0.161 & 0.57 ± 0.271 & 0.561 ± 0.135 & 0.547 ± 0.281 \\ 
 &  &  & Theta & 0.492 ± 0.346 & 0.772 ± 0.101 & 0.743 ± 0.157 & \textbf{0.862} ± 0.186 & 0.689 ± 0.129 & 0.413 ± 0.425 \\ 
 &  &  & Alpha & \textbf{0.852} ± 0.124 & 0.807 ± 0.131 & 0.861 ± 0.140 & 0.778 ± 0.167 & 0.779 ± 0.142 & 0.773 ± 0.244 \\ 
 &  &  & Beta & 0.844 ± 0.261 & \textbf{0.842} ± 0.098 & \textbf{0.907} ± 0.161 & 0.822 ± 0.145 & \textbf{0.807} ± 0.135 & \textbf{0.787} ± 0.376 \\ 
\midrule
\bottomrule
\end{tabular}
}
\end{table*}

We tuned the following hyperparameters in our study:
\begin{itemize}
    \item \textbf{Dropout rate}: Varied from $0.1$ to $0.5$. This parameter controls the probability of zeroing out elements in the dropout layers, which helps mitigate overfitting.
    \item \textbf{ASAPooling ratio}: Configured to vary between $0.1$ and $1.0$. It adjusts the node dimension reduction in pooling layers, affecting the structural summarization of the graph.
    \item \textbf{Number of Gated Graph Convolution (GGCN) layers}: Ranges from $1$ to $3$, influencing the depth of the graph network.
    \item \textbf{Number of output channels and layers in each GGCN}: Output channels are capped at $256$ with up to $17$ layers per GGCN. This configuration determines the network’s capacity and the complexity of each GGCN layer, influencing the learning and representation power of the network.
    \item \textbf{Learning rate}: Suggested over a logarithmic scale from $0.0001$ to $0.01$. This controls the size of the steps taken in the weight space, affecting the optimization speed.
    \item \textbf{Batch size}: Options include $16$, $32$, or $64$, determining the sample size for each forward and backward pass.
\end{itemize}

The evaluation of hyperparameter optimization was conducted using the Pareto front, taking into consideration a range of metrics such as Area Under the ROC Curve score(AUC), Precision, Specificity, and Recall. We also reported the accuracy and F1 score for each band in the result. The total number of parameters used for training depends on the depth of the model.

\section{Results and Discussion}
\subsection{Performance}
As observed, the AUC for the SVM baseline models ranges from 0.42 to 0.55, 0.42 to 0.70, and 0.41 to 0.85 for very mild, mild, and moderate to severe dementia, respectively, in distinguishing them from the control group. Notably, the variance of specificity is quite high.

However, we could see that our flexible GGCN approach, powered by Optuna's multi-subject hyperparameter tuning, has significantly improved model performance. As detailed in Table~\ref{tab:performance_comparison_0_1}, Table~\ref{tab:performance_comparison_0_2}, Table~\ref{tab:performance_comparison_0_3}, the dynamic GGCN model outperforms the baseline.
The analysis revealed that the alpha and beta frequency bands demonstrate superior outcomes in differentiating very mild dementia from the control group. Specifically, within the alpha band, using the Phase Locking Index (PLI) as the connectivity matrix yielded the highest overall performance in the first epoch, characterized by an AUC score of 0.97 (SD 0.07), F1 score of 0.84 (SD 0.16), precision score of 0.84 (SD 0.16), and recall score of 0.95 (SD 0.12). Although the specificity in the alpha band was relatively lower at 0.76 (SD 0.28), indicating variability, it still surpassed baseline model performance. 
In the case of the beta band utilizing the PLI, the results in the second and third epochs were notable, featuring an AUC of 0.91 (SD 0.13), F1 score of 0.90 (SD 0.16), precision of 0.94 (SD 0.14), recall of 0.89 (SD 0.16), and specificity of 0.91 (SD 0.22) for epoch 2, an AUC of 0.97 (SD 0.10), F1 score of 0.94 (SD 0.10), precision of 0.98 (SD 0.07), recall of 0.91 (SD 0.18), and specificity of 0.97 (SD 0.09) in epoch 3. 
From the result, we could see that in distinguishing mild dementia from the control group, the theta and alpha bands showed enhanced capabilities, however, it is important to note that no single frequency band consistently outperformed the others across all metrics, particularly exhibiting reduced efficacy compared to the differentiation of very mild dementia.
In contrast, in the result of distinguishing moderate to severe dementia, the theta bands exhibited optimal performance in all three epochs when utilizing the PLI. Furthermore, both the alpha and beta bands showed superior differentiation capabilities from the control group when Phase Locking Value (PLV) was used as the connectivity matrix with closely approaching 1 AUC scores. 
Overall, moderate to severe dementia exhibited the strongest results among all classifications of dementia severity.

\subsection{Pareto Front Hyperparameter Selection}

The optimal selection of hyperparameters was determined via the Pareto Front method, which evaluated four key metrics: AUC, Precision, Specificity, and Recall. Here, the AUC metric assesses the model's capacity to differentiate between classes. Precision indicates the accuracy of correctly predicted AD subjects, while Specificity measures the proportion of correctly identified HC subjects. Recall quantifies the proportion of accurately identified AD subjects among all AD cases.

As previously described, we utilized the unsupervised Nearest Neighbors method to pinpoint the most critical connections for each node by choosing specific K values, thereby creating a sparse graph. We conducted 50 experiments for each K value, which ranged from 5 to 17 across each band, totaling 650 trials per band to determine the optimal parameters. The multi-objective Pareto Front approach effectively identifies hyperparameter settings that strike an optimal balance between these competing metrics. This method facilitates a balance, such as maintaining good Specificity while not overly compromising Recall, thereby enhancing the model to address both classes equally.

Our proposed framework is highly flexible, which means that the number of GGCN blocks, output channels, and sequence length are hyperparameters that need to be tuned. The number of parameters generated from each band varies accordingly. To provide a better understanding, we compiled three tables for each CDR Rating in the Appendix (see Figures~\ref{fig:hyperparameter_settings_very_mild}, \ref{fig:hyperparameter_settings_mild}, and \ref{fig:hyperparameter_settings_moderate_to_severe}) showing the hyperparameters for each model. From the table, we can see that the combination of k, the number of GGCN blocks, the number of channels in each GGCN block, the ASAP ratio, and the dropout rate are all different for different bands.

\subsection{Explainability}
\subsubsection{Embedded Adjacency Matrix}

We created a topology map and a heatmap to illustrate the channel connectivity within the HC and AD groups. In the heatmap, a higher embedded correlation coefficient between two channels in the averaged adjacency matrix indicates stronger connectivity. We observed that several channel connections showed a noticeable difference between the HC and AD groups in the frontal lobe.

\begin{figure*}
    \centering
    \includegraphics[width=0.9\textwidth]{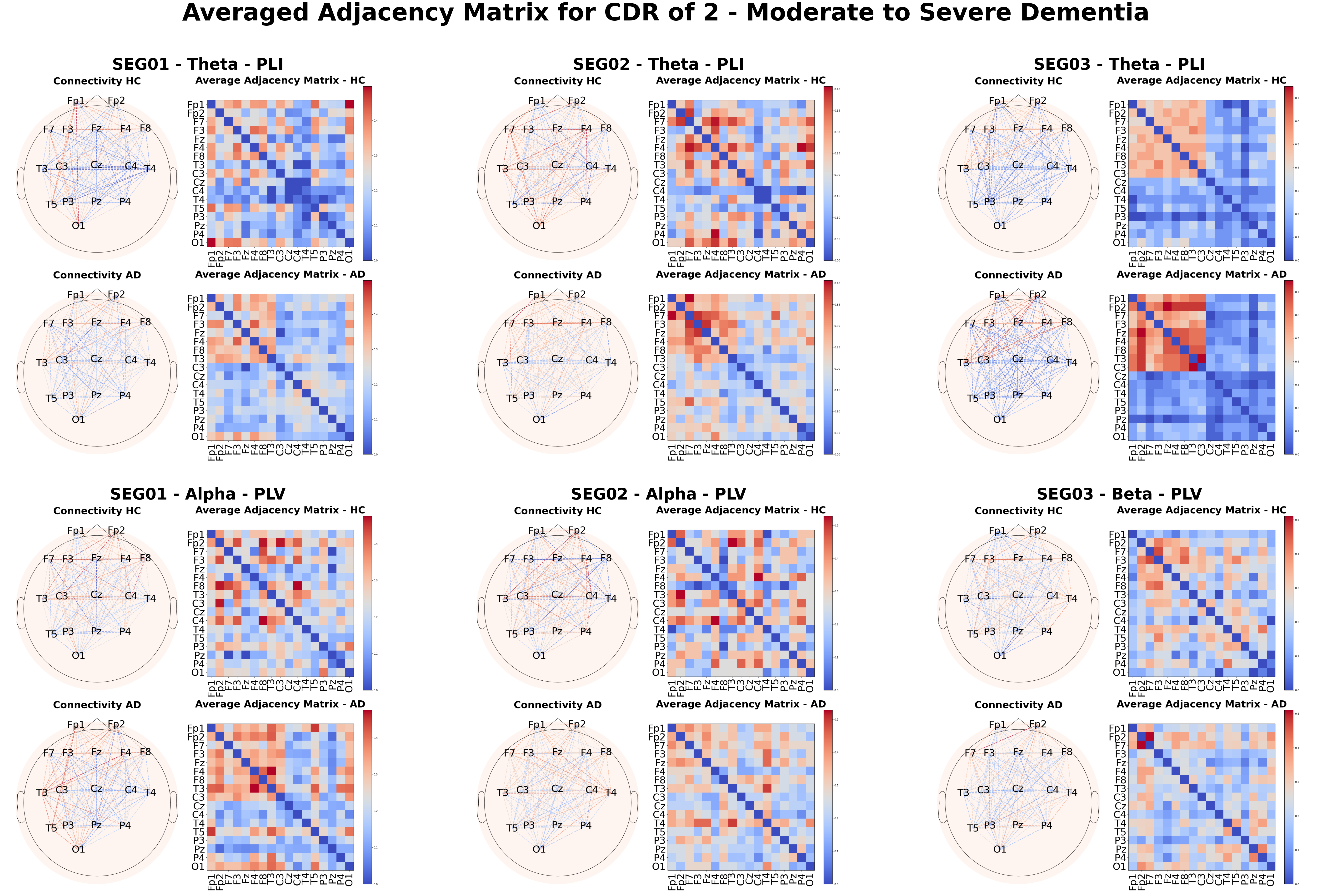}
    \caption{This figure shows the averaged adjacency matrix for the HC and AD groups in Moderate to Severe Dementia, highlighting differences between them.}
    \label{fig:moderate_to_severe_dementia}
\end{figure*}

For instance, the adjacency plot distinguishes the AD group with moderate to severe dementia from the HC group using Phase Locking Value (PLV) as a feature, as depicted in Figure~\ref{fig:moderate_to_severe_dementia}. In epoch 1, there is significant connectivity between C4-F8 (Right Primary Somatosensory Cortex in the Parietal Lobe to Right Frontal Lobe) and C4-Fp2 (Right Primary Somatosensory Cortex to Right Side of Prefrontal Cortex). In epoch 2, the connections between C4-F4 (Right Primary Somatosensory Cortex to Right Frontal Cortex) and C4-P4 (Right Primary Somatosensory Cortex to Right Parietal Lobe) are notable. Additionally, in epoch 3, a significant differentiation is observed in the connectivity between F3-F7 (Left Frontal Cortex to Left Frontal Cortex) and Fz-F3 (Midline Frontal Cortex to Left Frontal Cortex).

These findings suggest that EEG signals can effectively distinguish between AD and HC groups in the frontal and parietal lobe.

\section{Conclusion}

Alzheimer's Disease (AD) is a prevalent form of dementia that significantly impairs cognitive abilities such as thinking, acting, and reasoning, thereby severely affecting an individual's quality of life. Although the specific causes of AD vary among patients and are not entirely understood, early intervention is recognized as crucial for managing disease progression and enhancing the quality of life. To this end, modern approaches in machine learning, deep learning, and graph learning have been developed to facilitate early detection of AD.

EEG-based methods are particularly favored because they are non-invasive and can precisely capture brain signals via the scalp. However, the variability in EEG signals across individuals, due to anatomical and physiological differences, poses challenges in robustly and effectively distinguishing AD from healthy controls (HC).

In this paper, we analyzed EEG data from 123 subjects to classify them into HC and AD groups. We introduced a novel framework utilizing the Gated Graph Convolutional Network (GGCN) powered by the Multi-Objective Tree-structured Parzen Estimator (MOTPE) for hyperparameter optimization. The data were preprocessed, and Power Spectral Density (PSD) for each band was calculated for the Delta, Theta, Alpha, and Beta bands. 
We constructed the graph by calculating the PLI and PLV between channels and applying the Nearest Neighbors method for each band. This was followed by processing through one or several GGCN blocks followed by pooling to embed the features before sending into the classification layer.

This framework has demonstrated remarkable success in distinguishing HC from AD, especially in distinguishing HC and AD in the Moderate to Severe Dementia group, achieving AUC, Precision, Recall, Specificity, and Accuracy scores above 0.90, all with a lower standard deviation compared to other methods.

Moreover, we added explainability to our findings by examining the embedded adjacency matrix generated by our model. We visualized these differences directly on the scalp, where our analysis revealed a noticeable difference in connectivity in the frontal and parietal area of individuals with AD compared to healthy controls. Given the essential role of the Frontal Lobe in problem-solving, reasoning, and judgment, and the role of the Parietal Lobe in sensory perception and integration, these observations highlight the EEG biomarkers that may influence cognitive impairments in Alzheimer's Disease.

In the future, our objective is to enhance our framework to increase the specificity score while minimizing variability across different runs in distinguishing between HC and patients with mild and very mild dementia in Alzheimer's disease. We also aim to increase the speed of the model while maintaining or improving its overall performance score. Additionally, we are working on adapting this method to other datasets to enhance its generalizability.


%



\section*{Acknowledgment}
We would like to thank Dr. Albert C. Yang for generously providing the data used in this study. 
\bibliographystyle{IEEEtran}
\bibliography{IEEEbib}

\newpage
\onecolumn
\appendix
\section{Appendix}

\begin{figure}[!h]
    \centering
    \includegraphics[width=0.9\linewidth]{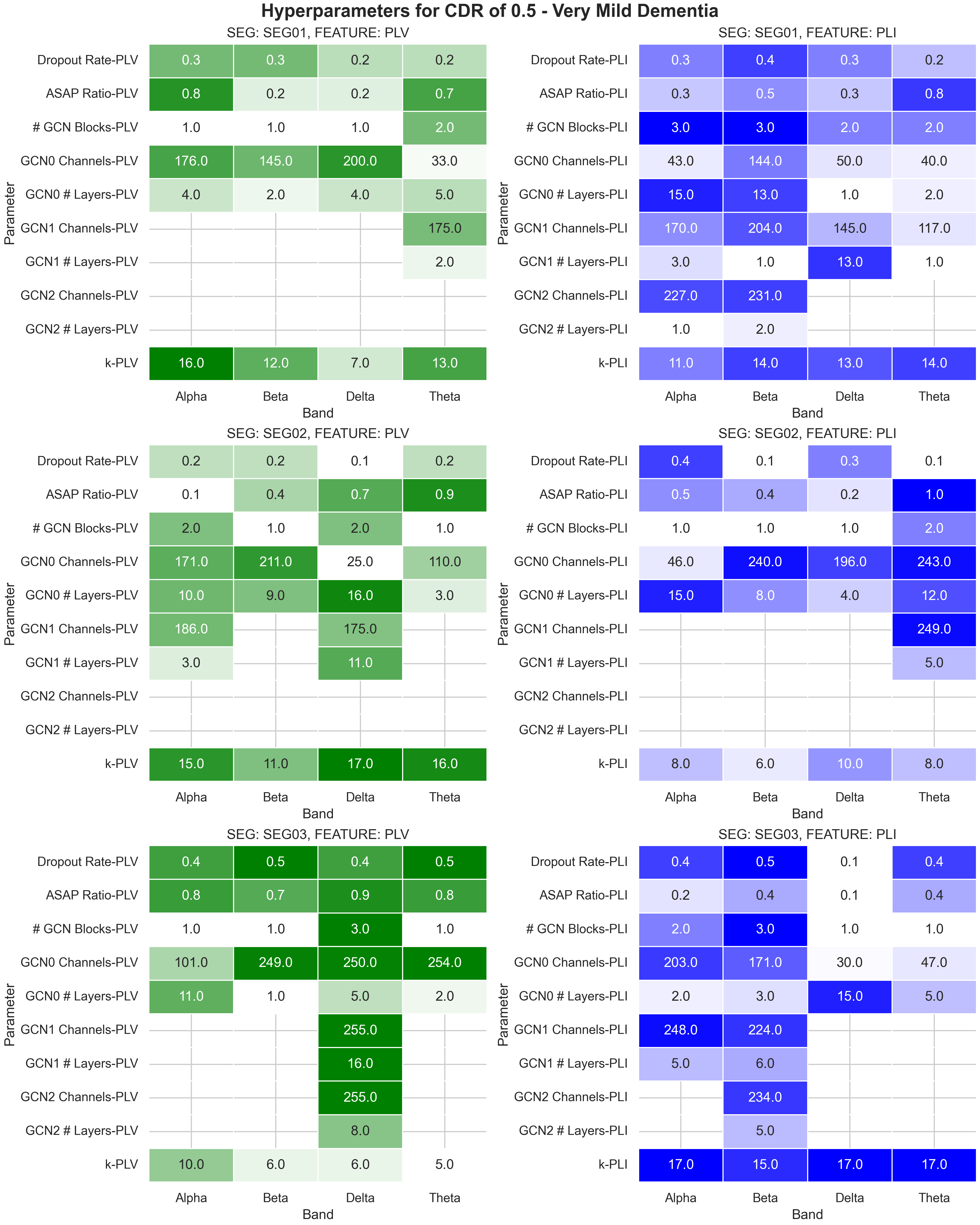}
    \caption{This figure shows the Hyperparameter Settings for CDR of 0.5 - Very Mild Dementia.}
    \label{fig:hyperparameter_settings_very_mild}
\end{figure}

\begin{figure}[!h]
    \centering
    \includegraphics[width=0.9\linewidth]{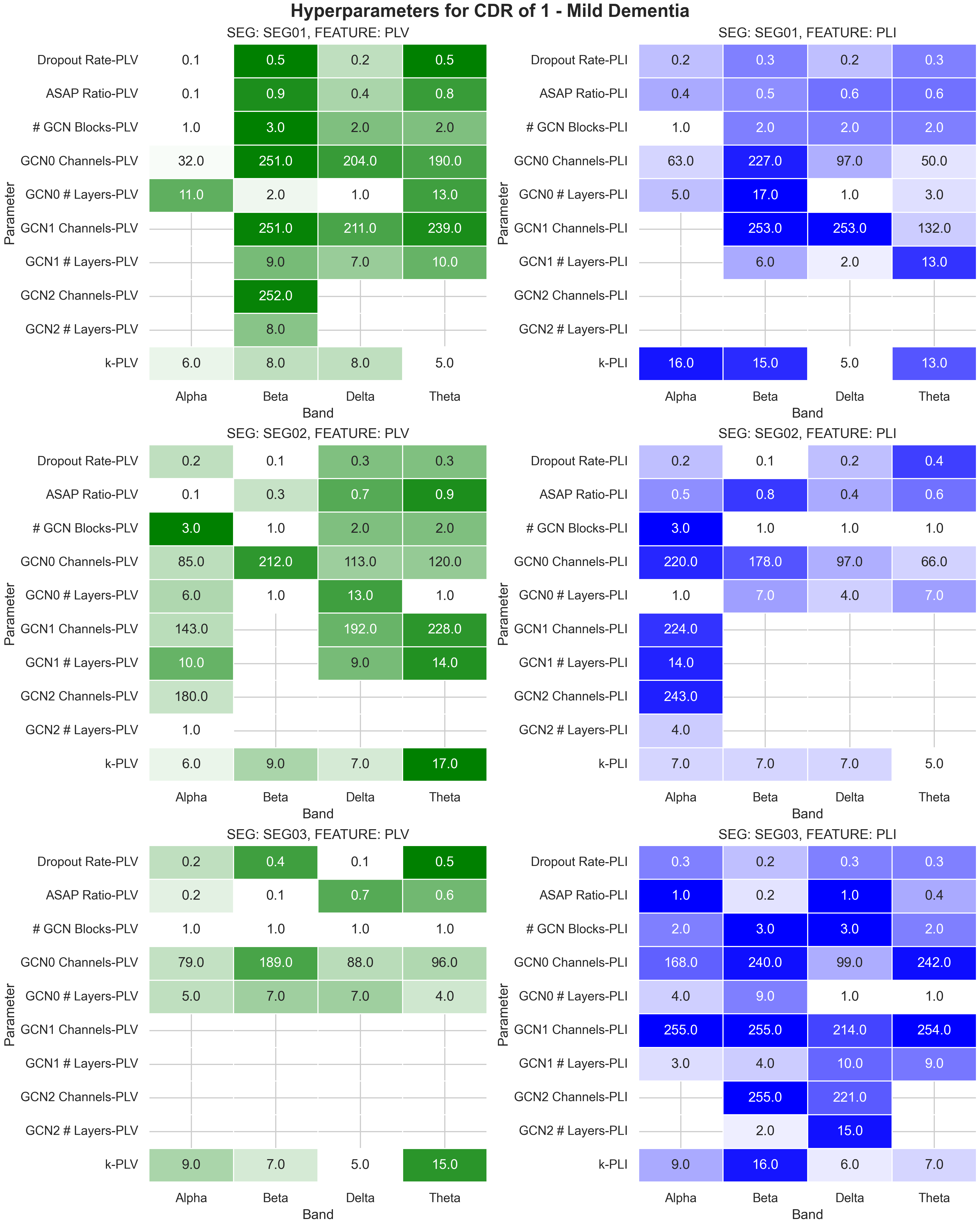}
    \caption{This figure shows the Hyperparameter Settings for CDR of 1 - Mild Dementia.}
    \label{fig:hyperparameter_settings_mild}
\end{figure}

\begin{figure}[!h]
    \centering
    \includegraphics[width=0.9\linewidth]{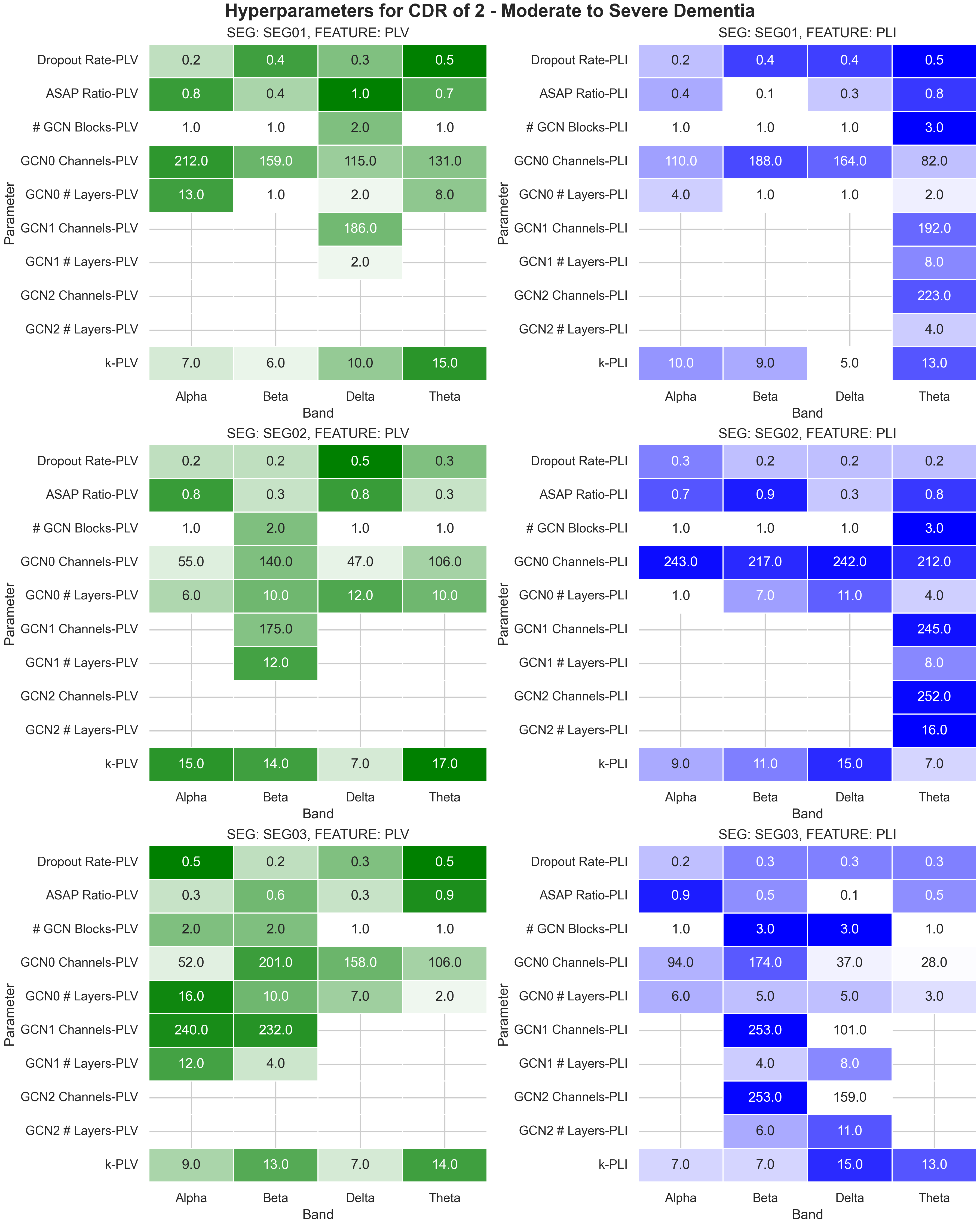}
    \caption{This figure shows the Hyperparameter Settings for CDR of 2 - Moderate to Severe Dementia.}
    \label{fig:hyperparameter_settings_moderate_to_severe}
\end{figure}

\begin{figure}[!h]
    \centering
    \includegraphics[width=0.8\linewidth]{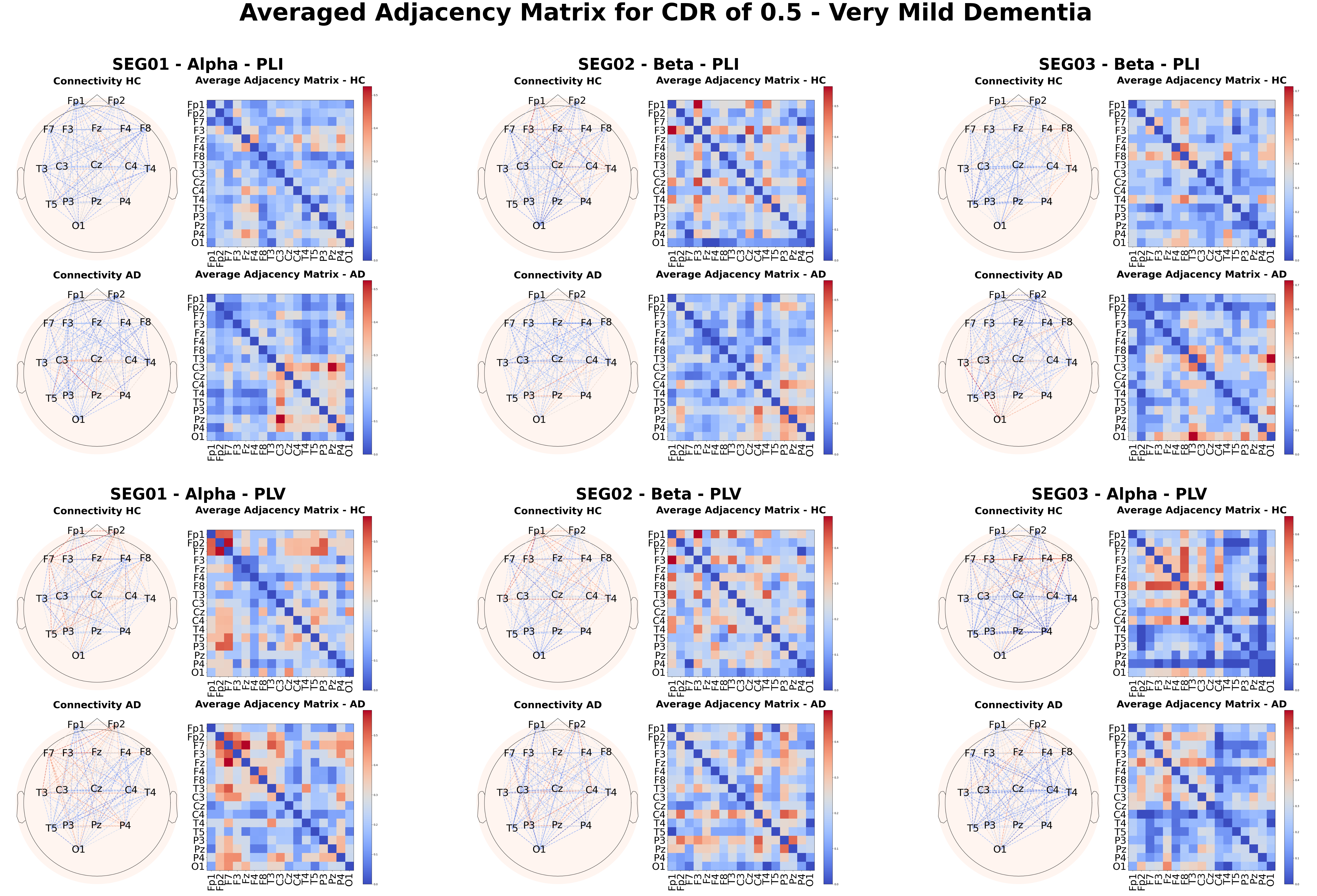}
    \caption{This figure shows the averaged adjacency matrix for the HC and AD groups in Very Mild Dementia, highlighting differences between them.}
    \label{fig:very_mild_dementia}
\end{figure}

\begin{figure}[!h]
    \centering
    \includegraphics[width=0.8\linewidth]{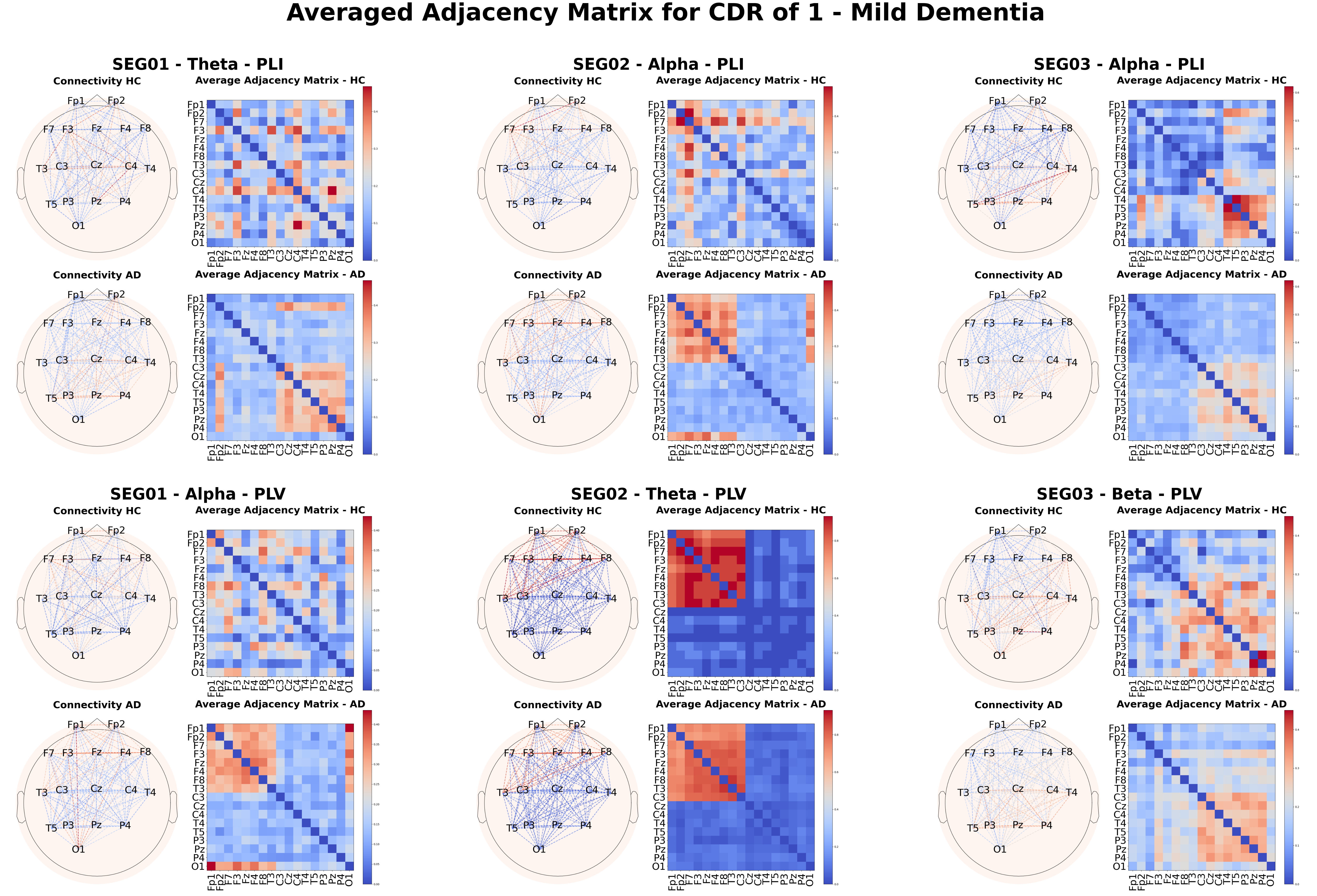}
    \caption{This figure shows the averaged adjacency matrix for the HC and AD groups in Mild Dementia, highlighting differences between them.}
    \label{fig:mild_dementia}
\end{figure}
\end{document}